\documentclass[10pt,twocolumn,letterpaper]{article}

\usepackage{cvpr}
\usepackage{times}
\usepackage{epsfig}
\usepackage{graphicx}
\usepackage{amsmath}
\usepackage{amssymb}

\usepackage{algorithm}
\usepackage[noend]{algpseudocode}
\usepackage{comment}
\usepackage{subcaption}
\usepackage{bm} % Bold math character.
\usepackage{cases}
\usepackage{booktabs} % For formal tables
\usepackage{diagbox}
\usepackage{makecell}
\usepackage{multirow}
\usepackage{array}

\usepackage[pagebackref=true,breaklinks=true,letterpaper=true,colorlinks,bookmarks=false]{hyperref}

\cvprfinalcopy % *** Uncomment this line for the final submission

\def\cvprPaperID{8845} % *** Enter the CVPR Paper ID here

\def\name{\texttt{GearNN}\xspace}
\newcommand{\rev}[1] {{#1}}

\newcolumntype{L}[1]{>{\centering\arraybackslash}p{#1}}

\ifcvprfinal\fi

\begin{document}

%%%%%%%%% TITLE
%\title{Heterogeneous Input Neural Network Adaptor}
\title{Partial Weight Adaptation for Robust DNN Inference}

\author{Xiufeng Xie\\
Hewlett Packard Labs\\
{\tt\small xiufeng.xie@hpe.com}
\and
Kyu-Han Kim\\
Hewlett Packard Labs\\
{\tt\small kyu-han.kim@hpe.com}
}

\maketitle
%\thispagestyle{empty}

%%%%%%%%% ABSTRACT
\begin{abstract}
%% 4000 characters limit
% invariant
% different distribution for training and inference
% the same distribution but too diverse
Mainstream video analytics uses a pre-trained DNN model with an assumption 
that inference input and training data follow the same probability distribution. 
However, this assumption does not always hold in the wild: autonomous 
vehicles may capture video with varying brightness; unstable wireless 
bandwidth calls for adaptive bitrate streaming of video; and, inference 
servers may serve inputs from heterogeneous IoT devices/cameras.
In such situations, the level of input distortion changes rapidly, thus
reshaping the probability distribution of the input. 
% flatness
% Even trained with data augmentation, a constant DNN fails to accommodate such diverse inputs.
% For example, a DNN must sacrifice the accuracy on high-quality inputs to guarantee the accuracy on low-quality inputs.  
%
% KH: due to space constraint
%One workaround is to fine-tune multiple DNNs, each for a particular input
%distortion level, and then switch between them based on the distortion level of
%the current input. However, this requires keeping multiple DNNs in memory
%to avoid the high latency to initialize and warmup DNNs, causing huge overhead.
% KH: cut ends here. 
% in either memory usage or latency. 
%, inference accuracy and training cost

We present \name, an adaptive inference architecture that accommodates
heterogeneous DNN inputs. \name employs an optimization algorithm to identify 
a small set of ``distortion-sensitive'' DNN parameters, given a memory budget. 
Based on the distortion level of the input, \name then adapts
only the distortion-sensitive parameters, while reusing the rest of DNN
parameters across all input qualities.  In our evaluation of DNN inference with
dynamic input distortions, \name improves the accuracy (mIoU) by an average of
18.12\% over a DNN trained with the undistorted dataset and 4.84\% over
stability training from Google, with only 1.8\% extra memory overhead. 
%\name architecture applies to any DNN model. 
%\name adapts a small portion of DNN weights following the category of current input. 
\end{abstract}

%%%%%%%%% BODY TEXT
\section{Introduction}
\label{sec:intro}

Video analytics solutions typically use a DNN with pre-trained weights
for inference, assuming consistent probability distribution between the training
and test dataset. Unfortunately, the inputs to DNN inference might have various 
distortions that alter the probability distribution and harm DNN performance in practice. 
An autonomous vehicle may drive in and out of shades, causing abrupt brightness 
change in the captured video; a drone may change a compression ratio of video frames 
while streaming to the inference server based on wireless link bandwidth; and
edge servers may need to process data from IoT devices with heterogeneous  
camera hardware and compression strategies. In these scenarios of input
distortions with high dynamic range, existing solutions that rely on a DNN with constant
pre-trained weights suffer from severe accuracy loss~\cite{stability}. We
observe up to a $58\%$ accuracy loss in our experiments (\S\ref{sec:expe}).
%% TODO: verify the above number later

%.~\footnote{A semantic segmentation model 
%trained with undistorted data suffers from a   
%when taking a compressed video input.}. 

%, like moving under a bridge or pointing to the sunshine after a sharp turn

\begin{figure}[t]
\begin{center}
\includegraphics[width=0.99\linewidth]{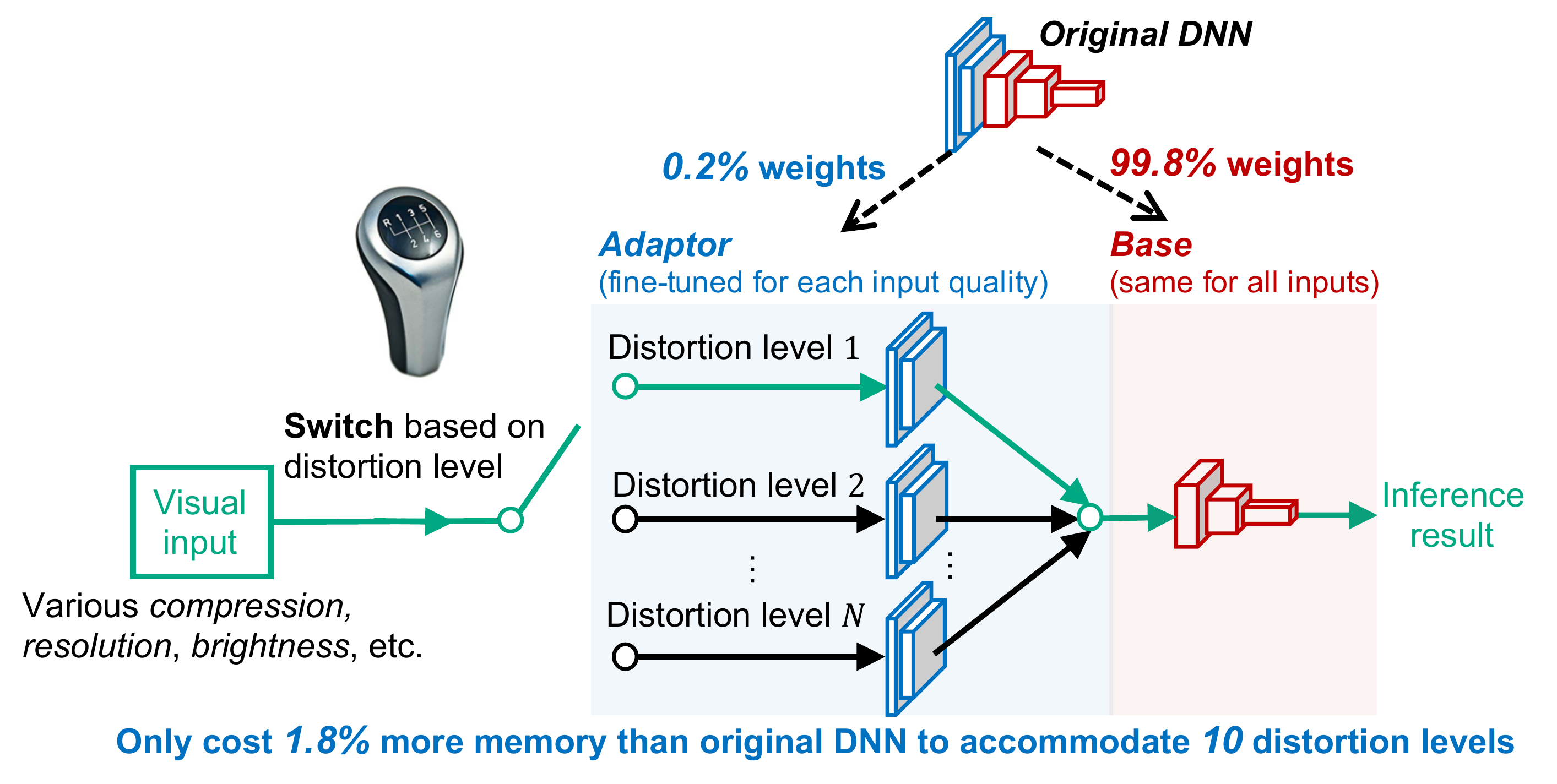}
\end{center}
\vspace{-5mm}
   \caption{\name, an adaptive inference architecture (This is a simplified illustration,
DNN layers in the \emph{adaptor} and those in the \emph{base} can be interleaved
with each other).}
\label{fig:outline}
\vspace{-5mm}
\end{figure}
%They do not have to be completely separated like in this simplified illustration

One workaround to handle input with unstable distortion level is to train a DNN 
for each possible distortion level by augmenting the training data to match that
particular distortion level, and then switch between them following the
distortion level of the current input. However, there are enormous distortion levels
(\eg, JPEG has 100 quality levels), running so many DNNs concurrently is
infeasible, due to limited memory.  Swapping DNNs between disk and memory causes 
huge latency, and thus, is impractical. 
%
%Data augmentation also fails to address this problem. Taking image compression
%as an example, images with lower quality have more noise in the high-frequency
%components. If the training dataset has mixed quality, the DNN learns not to
%look at the high-frequency components, which sacrifices the performance on
%high-quality inputs.  Therefore, it is difficult to use a single DNN to
%guarantee the inference performance for both high and low input qualities.

This paper proposes \name, an adaptive DNN inference architecture to accommodate
real-world inputs with various distortions, without sacrificing memory
efficiency.  \name only adapts a tiny portion (\eg, $0.2\%$ of the DNN size in
\S\ref{sec:impl}) of the DNN weights that are ``distortion sensitive''
(called \emph{adaptor}) to fit the distortion level of the instantaneous input,
while reusing the majority of weights (called \emph{base}) across all inputs.
In this way, the adaptation leads to high inference accuracy, while reusing most
weights guarantees the memory efficiency and scalability. We name our design
\name: like the gearbox that helps a single engine handle different car speeds, 
\name helps a single DNN base to accommodate a wide range of input distortions.

The \name workflow can be summarized as follows:

\emph{(i) Identifying distortion-sensitive weights offline.} 
Given a DNN pre-trained with undistorted training dataset, \name first
fine-tunes this DNN to multiple versions, each with training data of 
a particular distortion level. 
Next, by comparing the original DNN and the fine-tuned versions, \name runs an
optimization problem (\S\ref{sec:knapsack}) to identify a set of
distortion-sensitive DNN weights, under a constraint of additional memory
budget.

\emph{(ii) Partial DNN fine-tuning offline.} \name then partially fine-tunes the
DNN for each pre-defined distortion level, by only updating the
distortion-sensitive weights (\ie, \emph{adaptor}), while freezing the rest
of the pre-trained DNN weights (\ie, \emph{base}). This step yields 
multiple \emph{adaptors}, each for a particular distortion level.  

\emph{(iii) Partial DNN adaptation online.} With multiple fine-tuned small
\emph{adaptors} and a single copy of the \emph{base} loaded in memory, \name
switches between the \emph{adaptors}, following a current input distortion level
(like compression level), while reusing the \emph{base} across all
possible inputs. 

We have prototyped \name on top of popular DNN models like
\texttt{DRN}~\cite{drn1, drn2} and \texttt{Mask R-CNN}~\cite{mask} using \texttt{PyTorch}, and performed an extensive evaluation with semantic segmentation and detection
tasks (\S\ref{sec:expe}). 
Our evaluation shows that \name enables robust DNN inference under
various input distortion levels, while retaining memory efficiency. 
Specifically, \name achieves up to $18.12\%$ higher average inference accuracy
over the original DNN trained with undistorted data. \name also outperforms
other alternatives such as stability training from Google or fine-tuning a DNN
using a dataset with mixed distortion levels, and it consumes only $1.8\%$ more
memory over such single DNN solutions to accommodate $10$ distortion levels. 
%(across all distortion levels) 
Meanwhile, compared to switching between multiple DNNs, \name reduces memory
consumption by $88.7\%$, while achieving a similar level of accuracy.

%In principle, \name can be applied to enhance the quality-robustness of any DNN model. 
%We transform the original \texttt{PNG}-formatted dataset to multiple H.264 video copies, 
%each having a particular quality level (\ie, constant rate factor or CRF). We record
%the inference accuracy (mIoU) and memory consumption of the solution. We
%perform the same evaluation for \texttt{JPEG} images with quality levels ranging
%from 10 to 100.  Similarly, we convert the original \texttt{Cityscapes} dataset
%to multiple copies, each having a particular brightness level. 

%, using \texttt{Cityscapes}~\cite{cityscapes} dataset and two DNN models \texttt{DRN-D-22} and \texttt{DRN-D-38}~\cite{drn1,drn2}

% KH: we can reduce / remove this paragraph.. 
%It is worth noting that \name significantly saves training overhead in
%both energy and time. Since it partially fine-tunes a  
%pre-trained DNN, \name needs to run a small number of epochs (4 in our implementation). 
%In this way,  

Our contributions can be summarized as follows:
\begin{itemize}
\item We propose \name, a general technique to improve the tolerance of DNN
models to input distortions with high dynamic range, without compromising the memory efficiency.
We validate \name over state-of-the-art DNNs, where it outperforms existing solutions.
\item \name formulates an optimization problem that selects a tiny set of DNN
weights to be adapted, under a given constraint of memory consumption. 
% the frequency response as the nature of fine tuning
%
\item \name is the first to enable robust video analytics on adaptive bitrate
(ABR) streaming, following the trend of modern video streaming. 
\item \name can quickly customize (only takes 4 training epochs in our
prototype) any pre-trained DNN to fit the desired dynamic
range of input distortion, offering flexibility in deployment. 
\end{itemize}

\section{Related Work}
\label{sec:related}

\textbf{Adaptive neural network.}
Existing work~\cite{adaptivenn, skipnet} discusses adapting the neural network 
architecture based on the instantaneous input. However, these solutions focus only on
accelerating the inference by early exit~\cite{adaptivenn} or skipping
convolution layers~\cite{skipnet}. In contrast, \name aims to improve 
the DNN robustness under diverse \& dynamic input distortions without compromising
memory consumption. Besides, \name adapts DNN weight values, instead of
network architecture, which guarantees fast adaptation and backward compatibility. 

\begin{figure}[t]
\begin{center}
\includegraphics[width=0.90\linewidth]{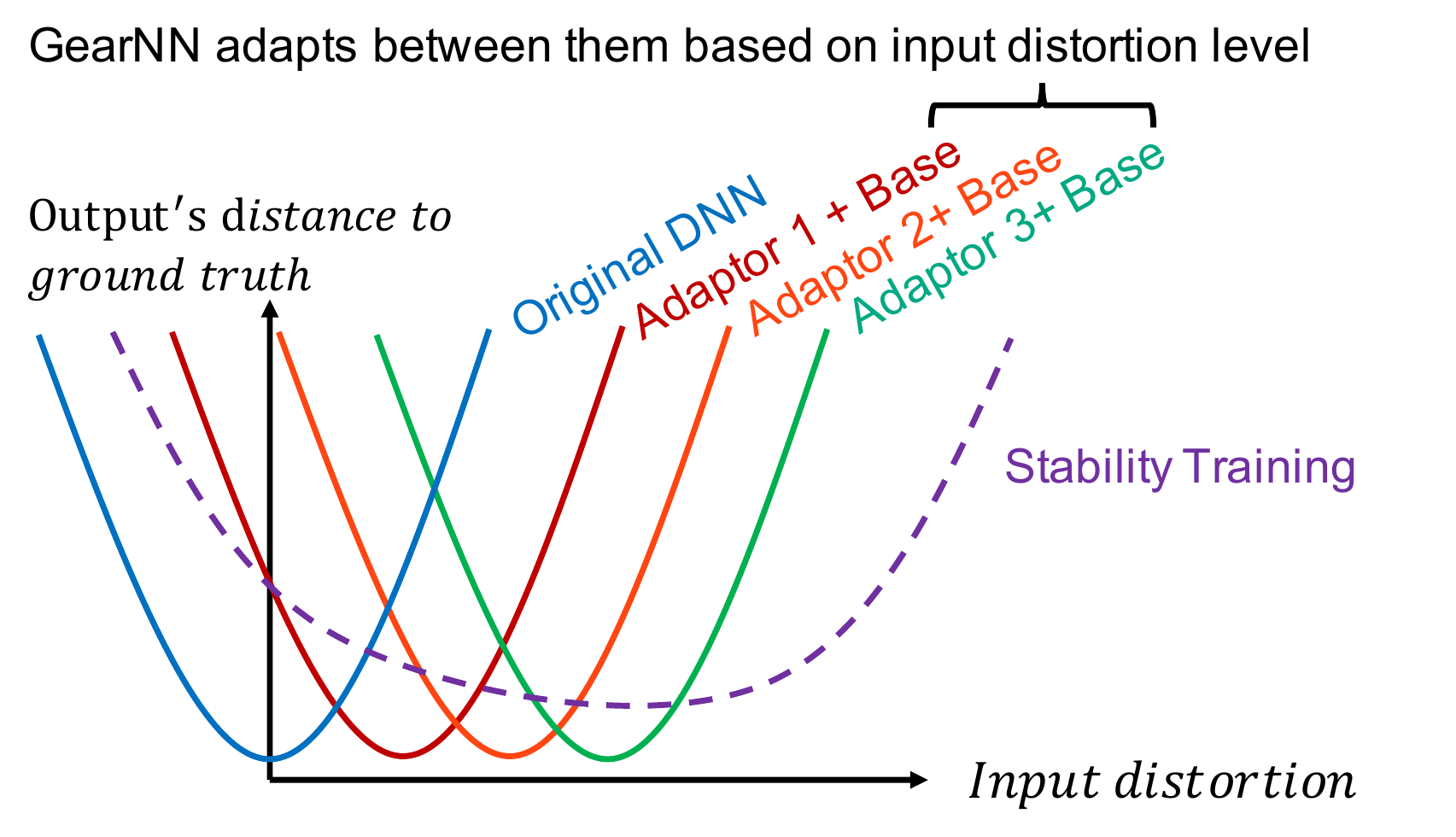}
\end{center}
\vspace{-5mm}
   \caption{An adaptive DNN can better serve a wide dynamic range of input
distortions than a constant DNN.}
\label{fig:no_single}
\vspace{-5mm}
\end{figure}

\textbf{Improve training for better robustness.} Some existing work proposes to
improve the training phase to make DNNs more robust to small input perturbations.  
\emph{Stability training}~\cite{stability} from Google uses a modified training
architecture that flattens the input-output mapping in a small neighborhood of
the input image. %It still trains one constant DNN for inference.
However, for perturbation with a broad range, it is difficult to
flatten one input-output mapping, as shown in Fig.~\ref{fig:no_single}. Hence,
\emph{stability training} fails to tolerate various input distortions. 
On the other hand, \name processes the input with different distortion levels 
by using different input-output mappings, and thus, it is more tolerable
under various input distortions.  
%
%In contrast, \name navigates inputs with different qualities to different
%DNNs, while reusing most of the weights across these DNNs to retain
%memory efficiency.  
%to make a DNN produce similar outputs for inputs with and without small perturbations,

\textbf{DNN pruning.}
DNN pruning~\cite{prune1, prune2, prune3} identifies and removes trivial neurons 
in order to reduce the model size and computational overhead. This approach
looks similar to our solution at first glance, but has fundamental differences. 
\name focuses on the ``distortion-sensitive'' weights rather than the neurons 
that are important to the output. Some DNN weights can be insensitive 
to the input distortion, but can be vital to the output.  In other words, 
\name can reuse such weights across different input distortion levels, 
but the DNN cannot work correctly after pruning such weights.
Besides, DNN pruning uses a pruned constant DNN for inference, while \name
adapts (part of) DNN online during the inference.  

\textbf{Transfer learning.} 
Due to the high cost to build a large training dataset, transfer
learning~\cite{transfer1, transfer2, transfer3} is proposed to customize a
pre-trained model on one task to fit another task. 
%
%Instead of training the DNN from scratch with randomly initialized weights, we
%can first train a DNN using a large general-purpose dataset, and then use
%it as a fixed feature extractor or start a new training process for the specific
%task with the initial weights from this pre-trained DNN.
%
Although \name follows the principle of transfer learning -- customizing the
pre-trained model to fit another task of same input type but different
distortion levels -- it is drastically different from existing practice of
transfer learning: \emph{(i)} \name actively identifies the
``distortion-sensitive'' weights and only fine-tunes such weights. In
contrast, typical transfer learning fine-tunes either only the fully-connected
layer or all DNN weights.
\emph{(ii)} \name enables partial weight adaptation in runtime inference,
whereas typical transfer learning still uses a pre-trained DNN for inference.

\textbf{Adaptive bitrate streaming.}
Adaptive bitrate (ABR) streaming quickly gains popularity in recent
years~\cite{bba, hls, dash}. By adapting the media stream quality in real-time
following the dynamic network bandwidth, ABR streaming achieves both the
high throughput and low latency. Unfortunately, existing DNN-based inference
solutions mostly use a single DNN and fail to accommodate the dynamic
quality levels on the ABR streams. In contrast, \name can adapt the DNN 
to accommodate dynamic streaming quality.

%\textbf{Adversarial Examples.}

%\textbf{Data augmentation and dropout.} A common technique to avoid over-fitting
%in DNN training is to perturb the training images by random rotation, cropping,
%or scaling. Dropout is another technique to combat over-fitting by randomly
%shutting down neurons during training. However, DNNs trained with such techniques
%still fail to handle the various inputs from IoT devices which suffer from much
%higher perturbation than that added during the training.

%\textbf{Training multiple copies of DNNs.} A straightforward solution to handle
%inference inputs with distinct qualities is to train multiple DNNs, each for a
%particular input quality. Due to the overhead of loading and warming up the
%DNN model (which costs more time than running inference on an image), the edge
%server needs to keep all DNNs in its memory. Unfortunately, keeping multiple DNNs in the memory damages the system scalability.  
%
%, then switch between them based on current input quality
%wastes the limited server memory and

\section{\name: Partial weight adaptation}
\label{sec:design}

In this section, we first analyze the effect of fine-tuning a DNN with
distorted training data (\S\ref{sec:design_1}), and then reveal our
observation that the majority of DNN weights have {\em minor} changes 
after fine-tuning with distorted data (\S\ref{sec:design_2}). Finally, we
elaborate on the \name design, featuring partial DNN adaptation 
motivated by our observation (\S\ref{sec:design_3}).      

\subsection{What happens when fine-tuning a DNN?}
\label{sec:design_1}

\subsubsection{Frequency-domain features of visual distortions}
\label{sec:distortions}
%The nature of reducing image quality is to add more noise. Furthermore, the noise is in the high-frequency band.
\rev{Since human eyes are insensitive to the high-frequency components (in the
spatial-frequency domain) of the visual input, encoders like JPEG and H.264
compress such high-freq. components aggressively.} 
Similarly, after reducing the image brightness, human eyes still perceive
object outlines (low-freq. components), but not textures (high-freq.
components).
In other words, typical visual distortions in practice essentially add noise
to the high-freq. components. 
%After being fine-tuned with a distorted dataset, the DNN should be able to cope with such high-frequency noise.   

\subsubsection{DNN's frequency response after fine-tuning}
\label{sec:response}

%As discussed in \S\ref{sec:distortions}, the practical visual distortions are
%equivalent to adding high-frequency noise to the visual input. In order to
%understand how DNN fine-tuning reacts to such high-frequency noise, we measure
%DNNs' response to the frequency-domain components of the input. 

\begin{figure}[t]
\begin{center}
\includegraphics[width=0.95\linewidth]{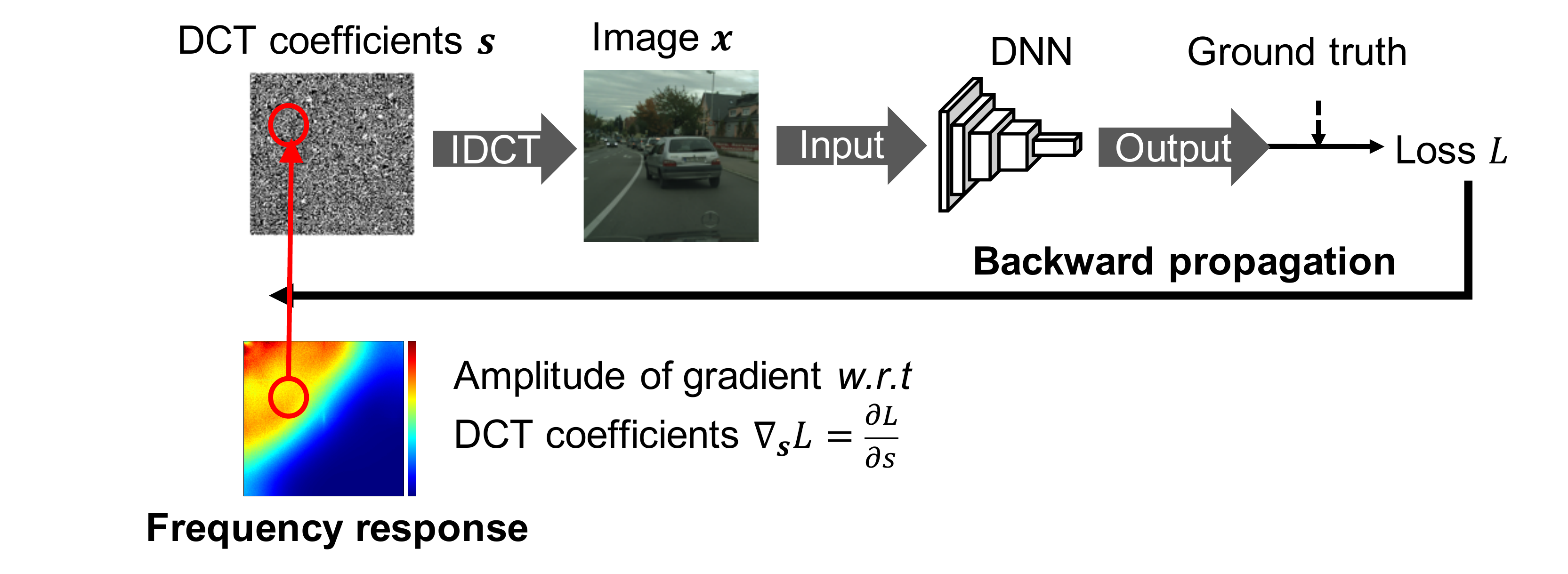}
\end{center}
\vspace{-7mm}
   \caption{Using the gradient of loss to model the DNN's freq.-domain
perceptual sensitivity (\emph{frequency response}).}
\label{fig:backprop}
\vspace{-5mm}
\end{figure}

%As discussed in \S\ref{sec:distortions}, typical visual distortions in practice
%are equivalent to adding high-frequency noise to the visual input. 
%
\rev{Inspired by human eyes' spatial frequency response, we measure DNNs' spatial
frequency response to understand how DNNs respond to the noise
caused by visual distortions (\S\ref{sec:distortions}). 
%Following the gradient-based DNN sensitivity modeling technique from
%, 
As shown in Fig.~\ref{fig:backprop}, we prepend an Inverse Discrete Fourier
Transform (IDCT) module to the typical DNN training framework, then the
frequency components (DCT coefficients) of the input become the leaf nodes, and
we can perform backward propagation to obtain the gradient of loss
\textit{w.r.t.} every input frequency component. A higher gradient amplitude of
a certain frequency component indicates that the DNN is more sensitive to input noise
on this frequency, and we define the gradient map of all input frequency
components as the DNN's \emph{frequency response}~\cite{grace}.}   

\begin{figure*}[t]
\begin{center}
\includegraphics[width=0.90\linewidth]{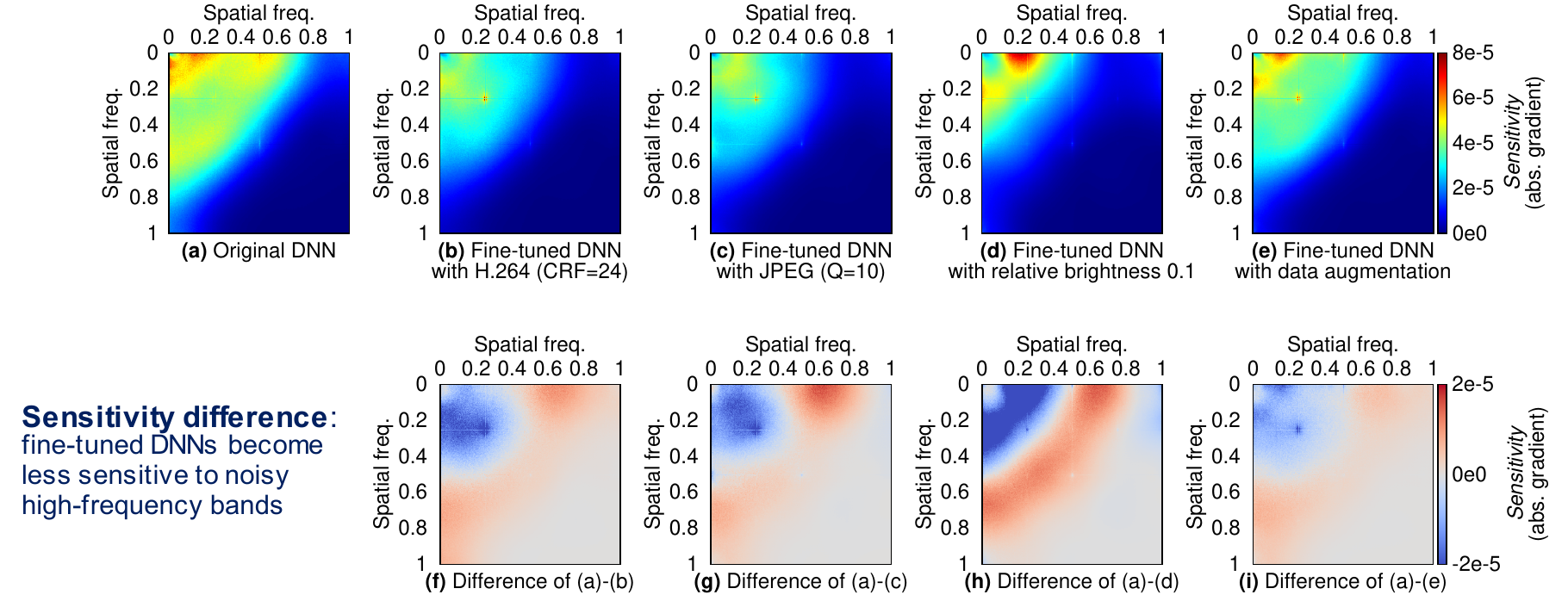}
\end{center}
\vspace{-5mm}
   \caption{Comparing the DCT spectral sensitivity (\emph{frequency response}) of the original and
fine-tuned \texttt{DRN-D-38}. We use different color palettes for
the sensitivity in the 1st row and the sensitivity difference (can be negative) in the 2nd row.}
\label{fig:all_sensitivity}
\vspace{-5mm}
\end{figure*}

We then compare the \emph{frequency responses} of the original DNN and the DNN
fine-tuned with distorted training data. 
%(more details in \S\ref{sec:quality})
We repeat the comparison for 4 types of distortions:
\emph{(i)} H.264 compression with a quality (CRF) of 24; \emph{(ii)} JPEG
compression with a quality of 10; \emph{(iii)} Underexposure with 10\%
brightness of the original image; \emph{(iv)} Data augmentation that mixes H.264
frames with different qualities (CRF from 15 to 24) in one training set.    
Fig.~\ref{fig:all_sensitivity} shows the measured DNN frequency responses. We
see that, compared to the original DNN, the DNNs fine-tuned with distorted
training data become less sensitive to the high-frequency components in all
tested cases, \ie, they learn to avoid looking at the noisy high-frequency
components for better robustness.  

%Since distortions in our cases are equivalent to adding noise to the high-frequency bands (\S\ref{sec:distortions}), t

\subsection{Distortion-sensitive DNN weights}
\label{sec:design_2}

%\begin{figure*}[t]
%\begin{center}
%\includegraphics[width=0.99\linewidth]{figures/all_layers.pdf}
%\end{center}
%\vspace{-5mm}
%   \caption{DNN fine-tuning leads to parameter changes. Some layers (like the
%bias) change more than other layers, and the layers containing the majority of
%the DNN parameters do not change much.}
%\vspace{-5mm}
%\label{fig_all_layers_change}
%\end{figure*}

\begin{figure*}[t]
    \centering
    \begin{subfigure}[t]{0.32\linewidth}
	\includegraphics[width=\textwidth]{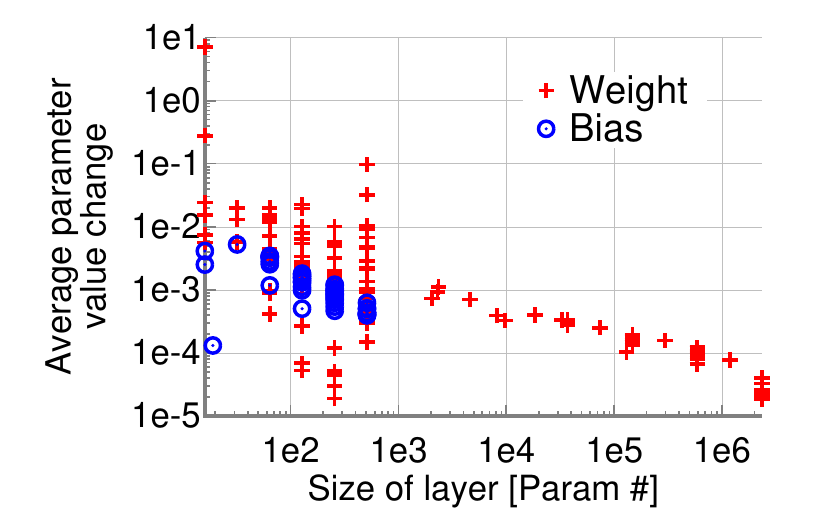}
	\vspace{-6mm}
	\caption{\texttt{DRN-D-38} fine-tuned with H.264 CRF=23.}
	\label{fig:change_drn_video}
	\end{subfigure}
	\hspace{3pt}
    \begin{subfigure}[t]{0.32\linewidth}
	\includegraphics[width=\textwidth]{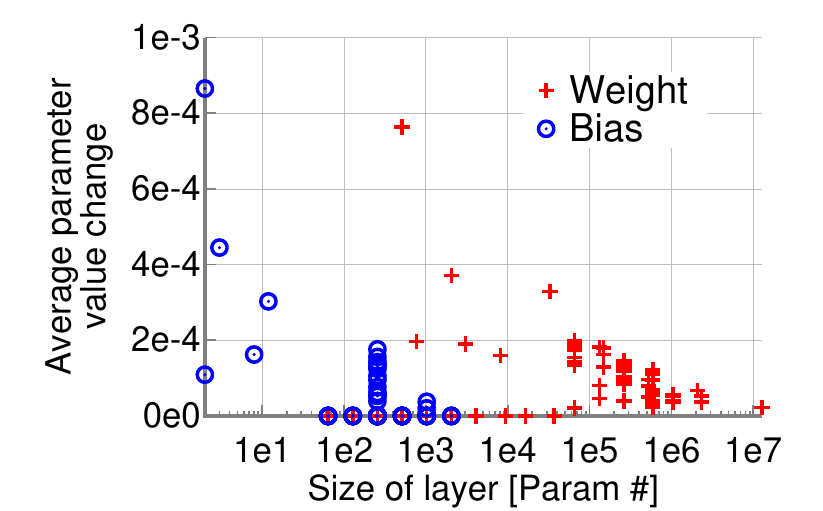}
	\vspace{-6mm}
	\caption{\texttt{Mask R-CNN} fine-tuned with JPEG Q=10.}
	\label{fig:change_maskrcnn_jpeg}
	\end{subfigure}
	\hspace{3pt}
    \begin{subfigure}[t]{0.32\linewidth}
        \includegraphics[width=\textwidth]{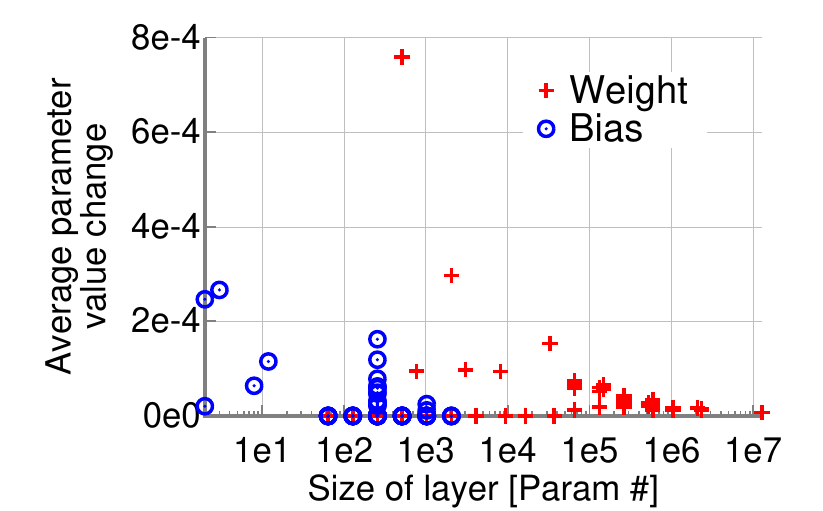}
        \vspace{-6mm}
        \caption{\texttt{Mask R-CNN} fine-tuned with relative brightness 0.2.}
	\label{fig:change_maskrcnn_dim}
    \end{subfigure}
        \vspace{-3mm}
	\caption{Per-layer average weight value change caused by fine-tuning.}
    \label{fig:change_amount}
        \vspace{-5mm}
\end{figure*}

Fine-tuning a DNN is updating its weights to fit a new training set.
In \S\ref{sec:design_1}, we reveal that fine-tuning a DNN with a distorted
dataset is equivalent to reshaping its \emph{frequency response} so that it
filters out the noise from the distortion. In this section, we dig further to
understand which DNN weights play more critical roles than other ones
in reshaping the \emph{frequency response}. We define such important weights
as ``distortion-sensitive''. 
It is worth noting that the ``distortion-sensitivity'' is different from the
importance of neurons in DNN pruning. Some weights can be vital for the
inference and should not be pruned. However, they can meanwhile be insensitive
to the input distortion, \ie, their values do not change much after fine-tuning 
with distorted training data -- there is no need to fine-tune them because the
inference output is insensitive to the small changes of the DNN weights as shown
in existing works~\cite{quantization1, quantization2, quantization3}.

\subsubsection{Modeling DNN weights' sensitivity to distortions}
\label{sec:changes}
% gradient is sensitivity
% changed amount is accumulated gradients
%
We first fine-tune a pre-trained DNN $\mathcal{D}$ by using distorted training datasets, and
compare the resulting weights with the original model.  
Let $K$ denote the number of layers in $\mathcal{D}$, and $\bm{l}_i$ denotes each
layer, which contains $N_i$ weights, then we have $\bm{l}_i = \{p_i^1, p_i^2
\ldots p_i^{N_i} \}$.
Let $p_i^j$ denote the $j$-th weight of the layer $\bm{l}_i$ in the original
pre-trained DNN, and $f_q(\cdot)$ denotes the fine-tuning process with a certain
distortion level $q$. Then, $f_q(p_i^j)$ is the corresponding weight of the
fine-tuned DNN, and we can compute the average change of weight values in layer
$\bm{l}_i$, caused by fine-tuning, as:  

\vspace{-5mm}
\begin{align}
\label{equ_vi}
v_i^q =
\frac{1}{N_i}\sum\nolimits_{j=1}^{N_i} \| p_i^j - f_q(p_i^j)\|
\end{align}
\vspace{-5mm}

%Based on the above analysis, we can conclude that the average weight value change $v_i^q$ of each layer captures its sensitivity to the visual input quality. 
%The gradient of loss \emph{w.r.t} DNN weights  
The layers with a high $v_i^q$ value yield significant change of weight values, 
when fine-tuned to fit the distortion level $q$, which means they are sensitive to 
the distortion level.  Therefore, we define $v_i^q$ as the \emph{distortion sensitivity} 
of layer $\bm{l}_i$. 
Following Eq.~\eqref{equ_vi}, we measure the layer-level weight change caused
by the fine-tuning for three cases: \texttt{DRN-D-38} fine-tuned with H.264 of
quality level (CRF) 23 in Fig~\ref{fig:change_drn_video}; \texttt{Mask R-CNN} fine-tuned with JPEG of quality
level (Q) 10 in Fig~\ref{fig:change_maskrcnn_jpeg}; and, \texttt{Mask R-CNN} fine-tuned with 
dimmed images of relative brightness 0.2 in Fig~\ref{fig:change_maskrcnn_dim}. 

In Fig.~\ref{fig:change_drn_video}, the sum size of the \texttt{DRN} ``weights''
(including both the weights \& biases)\footnote{We use the term ``weights'' to denote ``DNN parameters including both weights \&
biases'' for simplicity, except for Fig.~\ref{fig:change_amount}.} that changed more than $2\times10^{-4}$
after fine-tuning only accounts for 1.44\% of the model size. Similarly, only
$0.08\%$ and $0.0058\%$ of the \texttt{Mask R-CNN} weights changed more than $2\times10^{-4}$ in
Fig.~\ref{fig:change_maskrcnn_jpeg} and Fig.~\ref{fig:change_maskrcnn_dim}.  
We thus have an important observation: only a tiny portion of DNN weights
have non-negligible changes (\eg, $> 2\times10^{-4}$) after fine-tuning with
distorted data.
\emph{In order to fit DNN to the distorted inputs, we can simply reshape the
frequency response of a DNN by changing such tiny portion of DNN weights.}

\subsubsection{Choosing subset of DNN weights to fine-tune}
\label{sec:knapsack}

%, and we can leverage this finding in 
%selecting weights for memory-efficient fine-tuning. Further, the bias values 
%only account for $0.04\%$ of the total number of DNN parameters (in our test), 
%and we can keep multiple copies of the fine-tuned weights in the memory 
%with negligible memory cost and switches between them to fit the instantaneous 
%visual input quality. 
%%
%However, in practice, an inference server typically has more memory budget than
%keeping only the bias values. Meanwhile, fine-tuning more DNN parameters still
%leads to better inference performance on the new quality level.
Since only a tiny portion of DNN weights have non-negligible changes after
fine-tuning with distorted data, we propose to select and then fine-tune only these
``distortion-sensitive'' weights under a user-specified constraint
on memory space. Then, this can be formulated as a knapsack problem, where the
``item weight'' of layer $\bm{l}_i$ is its parameter number $N_i$, while the ``item
value'' of $\bm{l}_i$ is its \emph{distortion sensitivity} defined in
Eq.~\eqref{equ_vi}.
%its sensitivity to the input quality change, \ie, the
%average value of parameter change in $\bm{l}_i$ after fune-tuning with a
%particular input quality. Let $p_i^j$ denote the $j$-th parameter of the layer
%$\bm{l}_i$ in the original pre-trained DNN, $f_q(\cdot)$ denotes the fine-tuning
%process with a certain quality $q$, then $f_q(p_i^j)$ is the corresponding
%parameter of the fine-tuned DNN with a certain quality $q$, we define $v_i^q =
%\frac{1}{N_i}\sum_{j=1}^{N_i} \| p_i^j - f_q(p_i^j)\|$, which is the average
%changed amount of the parameters in layer $\bm{l}_i$ after fine-tuning.  
%of finding the adaptor $\mathcal{A}$ for quality level $q$
%
Now, the problem becomes selecting a list $S$ of DNN layers to maximize the total
``value'' (\emph{distortion sensitivity}) $\sum_{i \in S} v_i^q$, under the constraint
that the total ``size'' (memory consumption) $\sum_{i \in S} N_i$ is within a
user-defined bound $M$, which is summarized in Eq.~\eqref{equ_opt_s}. 
\vspace{-2mm}
\begin{align}
\label{equ_opt_s}
\max_{S} & \sum\nolimits_{\bm{i} \in {S}} v_i^q\nonumber\\
\textrm{s.t.} \quad & \sum\nolimits_{\bm{i} \in S} N_i \le M 
\end{align}
After obtaining the optimal list $S$ of layers, we fine-tune the corresponding
DNN portion $\mathcal{A}_q = \{l_i : i \in S \}$ (we call it \emph{adaptor}) with
the dataset of quality $q$ while keeping the rest of the DNN (we call it the
\emph{base} $\mathcal{B}_q=\mathcal{D}\setminus\mathcal{A}_q$) frozen. This
partial fine-tuning step can be denoted as $\mathcal{A}_{q}^{*} =
f_q(\mathcal{A}_q)$.

\subsubsection{Frequency response of partially fine-tuned DNN}
\label{sec:freq_validate}
\begin{figure}[t]
\begin{center}
\includegraphics[width=0.99\linewidth]{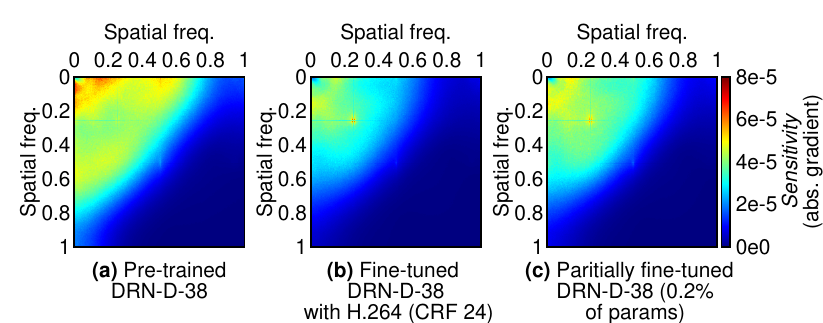}
\end{center}
\vspace{-7mm}
   \caption{Partial fine-tuning for \texttt{DRN}.}
\label{fig:partial_tuning_result}
\vspace{-4mm}
\end{figure}

\begin{figure}[t]
\begin{center}
\includegraphics[width=0.99\linewidth]{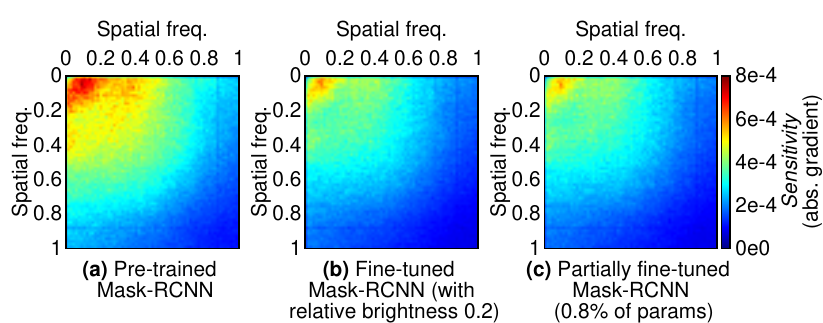}
\end{center}
\vspace{-7mm}
   \caption{Partial fine-tuning for \texttt{Mask R-CNN}.}
\label{fig:partial_tuning_result_mask}
\vspace{-1mm}
\end{figure}

%
%We then validate the idea of only fine-tuning a tiny portion of DNN parameters which are ``distortion-sensitive''.  % We fine-tune the DNN again, while updating only the bias values which changes more than the weights in our previous experiments-- all other DNN parameters get frozen. 
We further compare the frequency responses of both partially and fully fine-tuned
DNNs to understand the effect of partial fine-tuning.  
Our experiments include two models: \texttt{DRN-D-38} with $0.2\%$ of weights
fine-tuned (in Fig.~\ref{fig:partial_tuning_result}) and \texttt{Mask R-CNN} with
$0.8\%$ of weights fine-tuned (in Fig.~\ref{fig:partial_tuning_result_mask}).
By comparing Fig.~\ref{fig:partial_tuning_result}b and~\ref{fig:partial_tuning_result}c, we observe that
fine-tuning the entire \texttt{DRN} and fine-tuning only $0.2\%$ of the \texttt{DRN} weights
with the same distorted dataset yield DNNs with close frequency responses -- 
Both avoid looking at the noisy high-frequency components, compared to 
the original one in Fig.~\ref{fig:partial_tuning_result}a.
We have a similar observation from Fig.~\ref{fig:partial_tuning_result_mask}.

%The bias values have much higher $v_i^q$, and thus fine-tuning only the bias leads to a DNN with a similar frequency response to fine-tuning the entire DNN. 
%

% todo: show the frequency response retrain vs hinna

%\begin{figure*}[t]
%    \centering
%    %
%    \begin{minipage}[t]{0.48\linewidth}
%        \includegraphics[width=\textwidth]{figures/partial_tunning}
%        \vspace{-5mm}
%        \caption{Partial weight fine-tuning}
%	\label{fig:partial_tuning}
%    \end{minipage}
%	%
%	%\hspace{2pt}
%	%
%    \begin{minipage}[t]{0.48\linewidth}
%	\includegraphics[width=\textwidth]{figures/sensitivity_similar}
%	\vspace{-5mm}
%	\caption{Partial weight fine-tuning vs. entire DNN fine-tuning}
%	\label{fig:partial_tuning_result}
%	\end{minipage}
%\end{figure*}

\subsection{\name workflow}
\label{sec:design_3}
Based on the algorithm in \S\ref{sec:design_2}, we design \name, 
a robust inference architecture that adapts only a small portion 
of the DNN, following the instantaneous input distortion. 

%\subsubsection{\name workflow}
%\label{sec:workflow}

\textbf{Splitting the DNN (offline).}
%\label{sec:split}
The first step is to split the DNN into two parts -- the \emph{adaptor} to be
changed in real-time following the input distortion level and the
\emph{base} that remains constant.
% for all possible input qualities.
%
The splitting follows \S\ref{sec:knapsack}: 
For each distortion level $q$, \name fine-tunes the entire DNN using distorted
training data and compares the result with the original DNN, to obtain the
distortion-sensitivity metric $v_i^q$ of every DNN layer. Then, by solving the
optimization problem in Eq.~\eqref{equ_opt_s}, \name picks a subset of layers to
fine-tune, which is the \emph{adaptor} $\mathcal{A}_q$ for distortion level $q$.

\setlength{\textfloatsep}{1ex}
\begin{algorithm}[t]
\caption{DNN splitting \& partial fine-tuning (Offline).}
\label{alg:pwa}
{\small
\begin{algorithmic}[1]
    \Statex \textbf{Input:} $\quad\:$ $\mathcal{Q}=\{q_1,\ldots,q_N\}$ -- Supported distortion levels
     \Statex $\quad\:$ $\mathcal{D}$ -- Pre-trained DNN of $K$ layers, each having size $N_k$  
     \Statex $\quad\:$ $M$ -- Size constraint of each adaptor 
    \Statex \textbf{Output:} $\mathcal{A}^q$ -- Fine-tuned subset of DNN layers (\emph{adaptor})  
    %\State \textbf{/*Sensitivity evaluation on RGB channels*/}
    %\Statex \begin{center} \textbf{Partial weight fine-tuning (offline)} \end{center} 
    \ForAll{$q \in \mathcal{Q}$}  
    \State Fine-tune $\mathcal{D}_q^* = f_q(\mathcal{D})$ with the data of distortion level $q$   
    \State Compute $v_i^q = \frac{1}{N_i}\sum_{j=1}^{N_i} \| p_i^j -
f_q(p_i^j)\|$ for each $\bm{l}_i \in \mathcal{D}$   
    %\Statex
    \State Obtain ``values'' list $\bm{v}^q = \{v_1^q, \ldots, v_K^q \}$
using Eq.~\eqref{equ_vi}   
    \State Obtain ``weights'' list $\bm{n} = \{N_1,\ldots, N_K \}$   
    \State $S = \mathit{knapsack}(\mathit{value}=\bm{v}^q, \mathit{weight}=\bm{n}, \mathit{bound}=M)$   
    \State The subset of layers to fine-tune is $\mathcal{A}_q = \{\bm{l}_i : i \in S \}$   
    \State Partially fine-tune $\mathcal{A}_{q}^{*} = f_q(\mathcal{A}_q)$ %with data of distortion level $q$   
    \EndFor
    %\State \textbf{/*Selecting the DNN layers to fine-tune*/}
    %\State \textbf{/*Fine-tuning the selected DNN layers*/}
    %\Statex \begin{center} \textbf{Partial weight adaptation (online)} \end{center} 
    %\State Receive $T$, $W_{R}, W_{G}, W_{B}$ from the server  
    %\State Update $T$, $W_{R}, W_{G}, W_{B}$ of its codec to those from the server 
    %\State Perform online compression with the updated codec  
    %\State Always compute a window  $W_{cubic}$ of TCP Cubic   
%    \State On receiving a new $W_{claw}$   
%    \If {($\mathcal{L}==0$)}~/*LTE is the bottleneck*/ 
%	    \If {$N_s/N_t<1$} 
%		    \State /*Cellular-Informed BW Exploration*/
%		    \State $\texttt{cwnd} = {W_{claw}}$
%	    \Else 
%		    \State /*Hybrid BW Exploration*/
%		    \State $\texttt{cwnd} = \max{(W_{cubic}, W_{claw})}$
%	    \EndIf
%    \Else~/*Wireline bottleneck; fallback to PHY-informed Cubic*/
%    %\State 
%    %xyz: This is not explained anywhere
%    %xxf: fixed, should just fallback
%    \State $\texttt{cwnd} = W_{cubic}$%\frac{\mbox{TBS}_{i}}{\mbox{TBS}_{i-1}}$
%    \EndIf
\end{algorithmic}
}
\end{algorithm}

\begin{figure}[t]
\begin{center}
\includegraphics[width=0.99\linewidth]{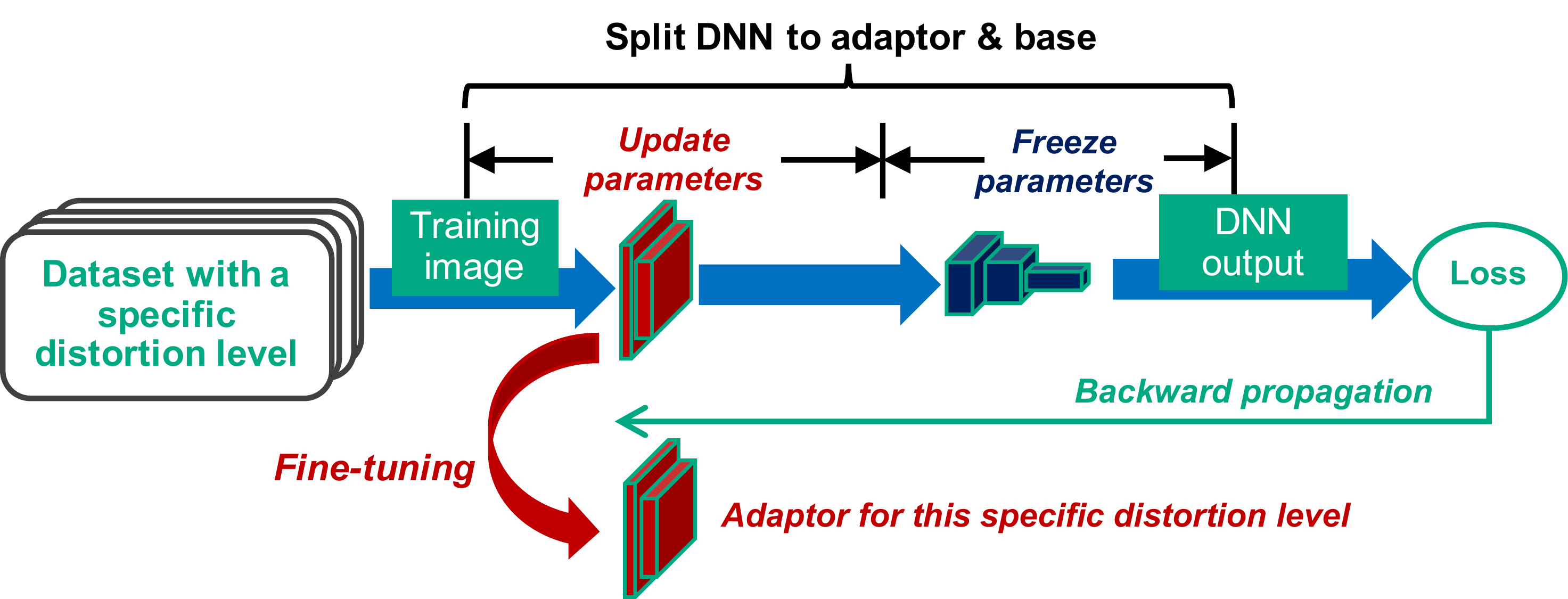}
\end{center}
\vspace{-6mm}
   \caption{Partial weight fine-tuning workflow.}
\label{fig:partial_tuning}
\vspace{-1mm}
\end{figure}

\textbf{Partial fine-tuning (offline).}
%\label{sec:fine_tune}
%
%After splitting the DNN to the \emph{adaptor} and \emph{base}, 
\name then fine-tunes the \emph{adaptors}, \rev{each using a dataset distorted
to a particular level (the orignal dataset before distortion is the same as the
one used in the DNN splitting step), while freezing the \emph{base},
as illustrated in Fig.~\ref{fig:partial_tuning}. 
In this way, \name obtains multiple fine-tuned \emph{adaptors} $\mathcal{A}_q^{*}$, each fitting a
particular input distortion level. 
Alg.~\ref{alg:pwa} summarizes the above steps.} 
%
%We define these parameter subsets as the \emph{adaptors} of our \name system,
%and the rest of the original DNN parameters pre-trained with the undistorted
%data as the \emph{base}.
%

\textbf{Partial adaptation (online)}
%\label{sec:adaptation}
%
With the fine-tuned \emph{adaptors} and the \emph{base}, we can run \name online. 
Since the size of adaptors is tiny, an inference server can load multiple
adaptors for all supported distorted levels at low memory cost. Next, given a
visual input stream with various distortion levels (\eg, ABR streaming or receiving
from heterogeneous IoT hardware), \name switches between the adaptors to fit
the instantaneous input distortion level, while keeping the base weights unchanged, 
as shown in Fig.~\ref{fig:outline}.
It is straightforward to get the distortion level of the instantaneous input
frame. % in the application scenarios discussed in this work. 
For JPEG images, the quality level is directly embedded in the image. 
For DASH video streaming, the frame resolution is obviously given. 
For the brightness levels, we can directly compute the brightness of an image 
based on its pixel values. Overall, it is simple and fast to determine which adaptor 
to use for the current input frame, enabling the real-time adaptation. 

\textbf{Why adapting at the layer-level?}
%\label{sec:layer_level}
%
%For adapting a subset of weights in a DNN model, we need not only their
\rev{Adapting a subset of DNN weights requires not only their new values but
also their positions in the model. For example, we can use a 0-1 vector to mark every
weight (1 means do adaption and 0 means otherwise), then a DNN with millions of
weights needs an extremely long vector, causing huge overhead.
In contrast, labeling the DNN layers incurs tiny overhead since DNNs typically have
up to hundreds of layers.} 

%However, labeling the positions such as using the keys in the parameter directory incurs extra costs. 

%For weight-level partial adaptation, we need to label the positions of every weight to adapt, which  overhead. 

\section{Implementation}
\label{sec:impl}

\subsection{DNN-related configurations}
\label{sec:dnns}
% drn and cityscapes
% ade20k and hrnet
% TODO: finalize with your final results
%From principle, \name works with any DNN model. We implement our solution on top of
%several different DNN models, including Dilated Residual Network, UperNet101,
%and HRNetV2. For the datasets, we evaluate \name on the Cityscapes and ADE20K
%datasets, two well-known datasets for semantic segmentation tasks. 
%
We implement \name using \texttt{PyTorch}~\cite{pytorch} and public datasets.  
While \name is designed as a generic inference architecture, 
%instead of a particular DNN model, 
we use two popular DNN models for its implementation: the dilated residual 
networks (\texttt{DRN})~\cite{drn1, drn2} and the \texttt{Mask R-CNN} model~\cite{mask_rcnn}. 
Also, we run \texttt{DRN} on the Cityscapes dataset~\cite{cityscapes} for
urban driving scenery semantic segmentation and \texttt{Mask R-CNN} on the Penn-Fudan
dataset~\cite{penn} for pedestrian segmentation and detection.
\rev{\name uses a small learning rate of $0.001$ and runs only $4$ epochs when
fine-tuning the adaptor as it only slightly updates the DNN trained with undistorted
dataset.}

Since \name allows a user to specify memory constraint, we evaluated \name
with various adaptor sizes, ranging from 0.1\% to 1\% of the original DNN size,
and our empirical results suggest that the adaptor with 0.2\% of the original \texttt{DRN}
size or 0.4\% of the \texttt{Mask R-CNN} size best balance the inference
accuracy and memory consumption. 

% KH-AAA
\subsection{Benchmark solutions}
\label{sec:bench}

For benchmarks, we use 4 alternatives (also in Fig.~\ref{fig:impl}):

%In what follows, we introduce the 4 benchmark solutions in more detail. 

\begin{figure}[t]
\begin{center}
\includegraphics[width=0.99\linewidth]{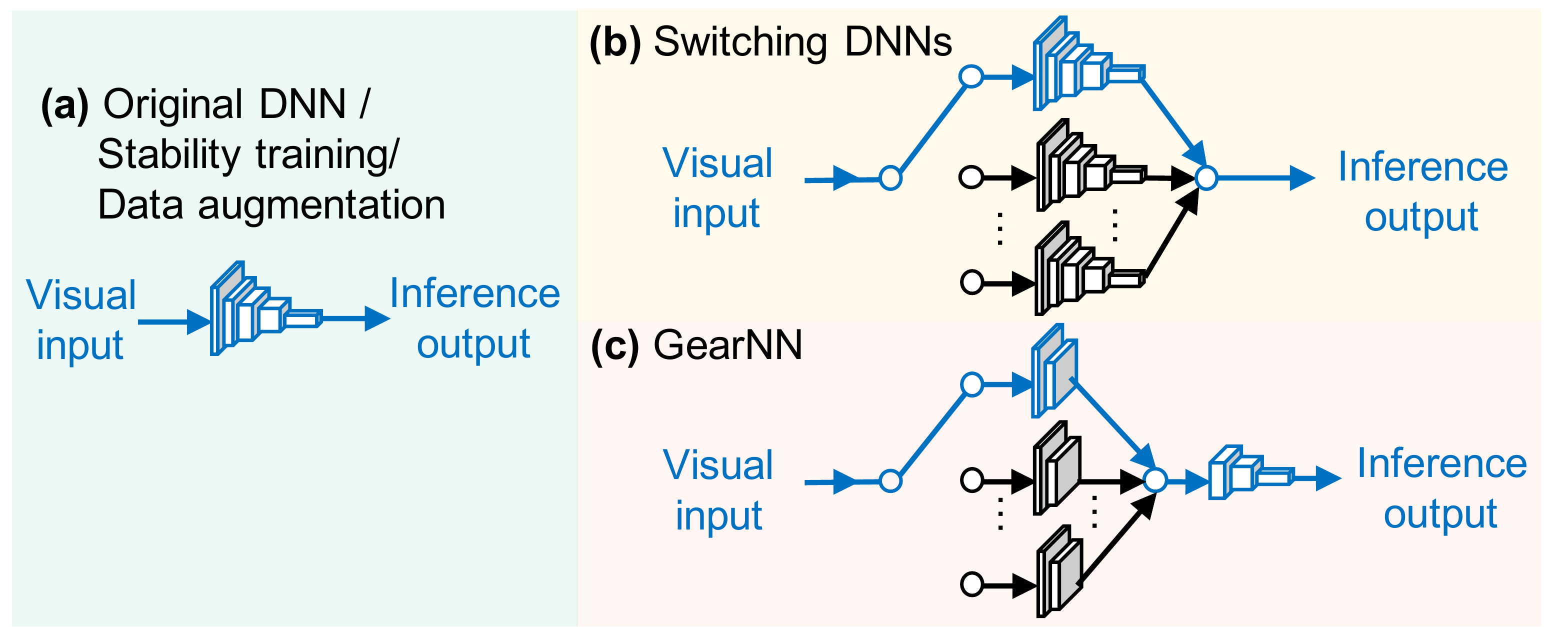}
\end{center}
\vspace{-5mm}
   \caption{Implementation of \name and benchmarks. Modules in blue are 
active when processing an input instance, while modules in black are
overhead.}
%\vspace{-2mm}
\label{fig:impl}
\end{figure}

\textbf{Original DNN.}
The original DNN is a DNN pre-trained with the original undistorted dataset and is available
in public. In our implementation, we download and use original DNN models 
from public links provided by their authors.

\textbf{DNN switching.}
A straightforward solution to accommodate DNN inputs with various distortions is
to train and use multiple DNNs, each for a particular distortion level.
%. We train and use multiple DNNs
%Due to the overhead of loading and warming up the DNN model (which costs more
%time than running inference on an image\cite{warmup}), the edge server needs to
%keep all DNNs in its memory.
%Unfortunately, keeping multiple DNNs in the memory damages the system scalability.

\textbf{Mixed training.}
A common technique to make DNN robust is to use perturbed images, via data 
augmentation, for training. We transform the original datasets to multiple versions, 
each for a particular distortion level, and then randomly sample from these
distorted datasets to form a new dataset with mixed distortions. Finally, we fine-tune 
the DNN with the mixed dataset.

\textbf{Stability training.}
Google proposed stability training~\cite{stability} to improve the robustness of
a DNN against noisy inputs. This solution forces a DNN to map inputs
with small perturbations to their original input images to achieve the same inference output. 
\emph{Stability training} only improves the training phase, and the
inference still uses a constant DNN.
%inference still uses a constant DNN with unmodified network architecture.
%However, it can only tolerate small perturbation and fails to handle the inputs from heterogeneous IoT devices.
%which essentially flattens the input-output mapping in a small neighborhood of the input image

\subsection{Distortion types}
\label{sec:quality}

We now define input distortion types used in \name.

%\textbf{\texttt{H.264} compression.}
%\texttt{H.264} video encoder controls the streaming rate (and the corresponding
%visual quality) by the \emph{constant rate factor} (CRF), whose value
%ranges from $0$ to $51$. The default CRF value is $23$, and a smaller CRF
%results in higher video quality. 

%\subsubsection{Downscaling resolution}
%\label{sec:dis_res}
\textbf{Resolution scaling.}
One primary motivation of \name is to make video analytics fit adaptive bitrate
(ABR) video streaming, and the standard way of ABR streaming is to adapt the
video resolution like MPEG-DASH~\cite{dash}. Downscaling the video
dimension reduces its size and causes distortions. We use H.264 encoder to
compress the Cityscapes dataset of original resolution
$\{2048\times1024\}$ to 6 different versions with smaller resolutions: 
$\{1920\times960$, $1600\times800$, $1280\times640$, $1024\times512$,
$800\times400$, $512\times256\}$. All these frames share a constant rate factor
(CRF) of 18, meaning small quantization loss.
% regardless of their resolutions.  

\textbf{JPEG compression.}
The quality of a JPEG image is defined by an integer value $Q$, ranging from $1$ to
$100$, and a higher $Q$ indicates higher image quality and less distortion. 
We compress a lossless PNG-formatted dataset (like Cityscapes) to 10
different versions with JPEG quality levels, ranging from $10$ to $100$ 
in a step of $10$, while keeping the same resolution as the undistorted dataset.
%~\cite{jpeg}

\textbf{Brightness.}
A too bright or too dark visual input can significantly affect the DNN inference
performance.
We adjust the image brightness by \texttt{Pillow} to make multiple versions of
the same dataset, each having a particular relative brightness to the original
dataset.  Our implementation includes relative brightness levels from 0.1 (10\%
of the original brightness) to 2.0 (double the original brightness), with the
resolution unchanged.

\rev{\name tackles distorted inputs by fine-tuning adaptors for each resolution,
compression level, brightness, or even combinations of them. In
\S\ref{sec:expe}, we evaluate different distortion types separately to show the impact of
each type.}

%%Then we compare the inference accuracy across these datasets.   
%\rev{\name can handle inputs with mixed distortion types, \eg, given $P$
%brightness levels and $Q$ resolutions, \name fine-tunes \& adapts between
%$P\times Q$ adaptors. In \S\ref{sec:expe}, we evaluate distortion types
%separately to show the impact of each type.}

%To show that \name can handle the brightness variation better than DNN trained with data augmentation, 

%\subsection{Memory constraint} % Configurations of our solution}
%\label{sec:config}
%% from 1% to 2%, we found 2% should be a good balancing point
%Since \name allows a user to specify memory constraint, we evaluated \name
%with various adaptor sizes, ranging from 0.1\% to 1\% of the original DNN size,
%and our empirical results suggest that the adaptor with 0.2\% of the original \texttt{DRN}
%size or 0.4\% of the \texttt{Mask R-CNN} size best balance the inference
%accuracy and memory consumption. 

%Unless otherwise mentioned, we use 0.2\% of the original DNN size as the memory constraint $M$ of the \name adaptor.   

\section{Experiments}
\label{sec:expe}
We now present experimental results under different distortion types, 
datasets, tasks, and benchmarks. %, to validate our \name design.

\begin{figure*}[t]
    \centering
    \begin{subfigure}[t]{0.32\linewidth}
	\includegraphics[width=\textwidth]{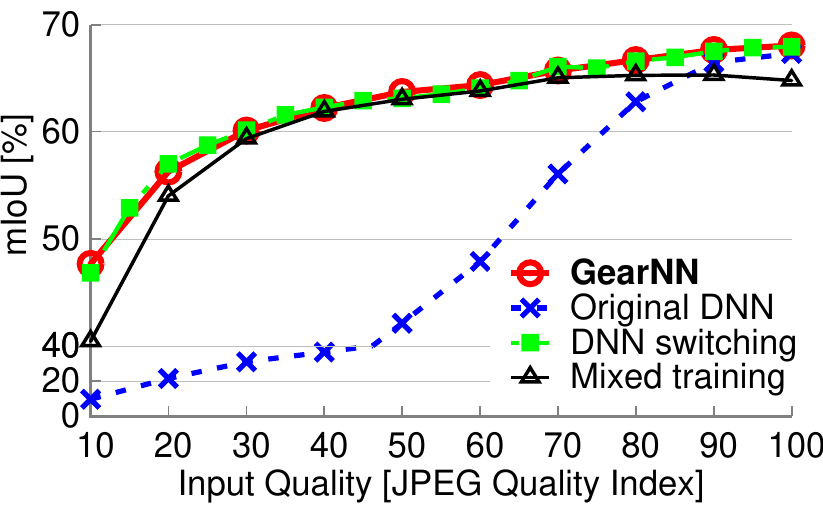}
	\vspace{-5mm}
	\caption{Different JPEG qualities.}
	\label{fig:accuracy_jpeg_22}
	\end{subfigure}
	\hspace{3pt}
    \begin{subfigure}[t]{0.32\linewidth}
        \includegraphics[width=\textwidth]{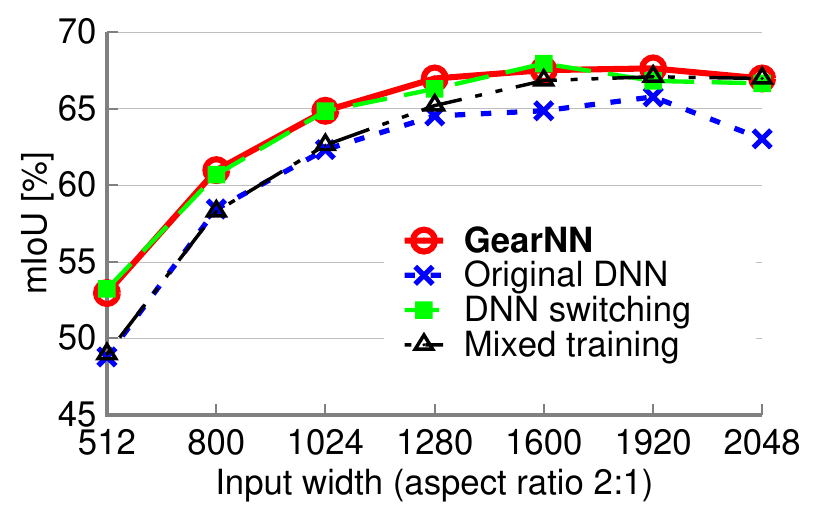}
        \vspace{-5mm}
        \caption{Different resolutions.}
	\label{fig:res_h264}
    \end{subfigure}
	\hspace{3pt}
    \begin{subfigure}[t]{0.32\linewidth}
	\includegraphics[width=\textwidth]{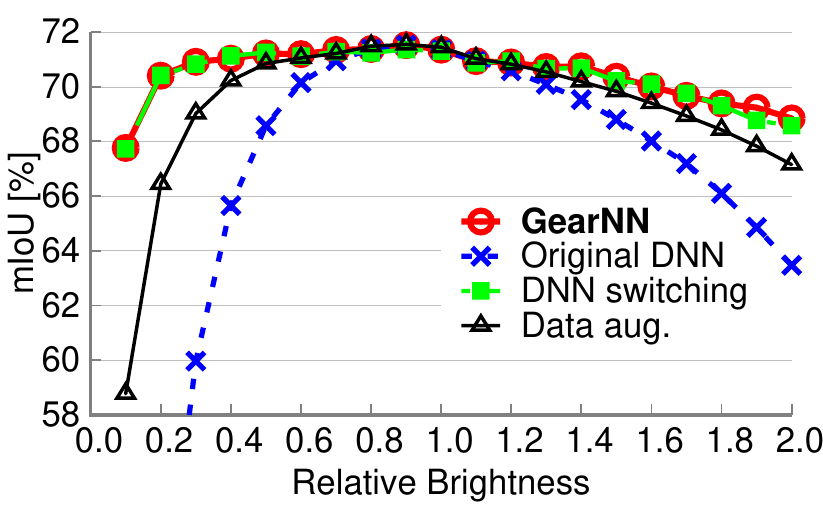}
	\vspace{-5mm}
	\caption{Different brightness.}
	\label{fig:bright_drn38}
	\end{subfigure}
	\vspace{-2mm}
	\caption{\name can accommodate various input distortion levels.}
\end{figure*}

\subsection{JPEG quality level}
\label{sec:jpeg}
%
%In a scenario where a server (at edge or cloud) processes the images 
%from remote IoT devices, JPEG is the dominant image compression solution. 
%The inference engine at the server might need to handle JPEG images 
%with various quality levels, while keeping accuracy and memory overhead under
%control.  In this context, we test \name with input of various JPEG quality
%, and it is critical to keep accuracy and memory overhead under control
In practical scenarios where a server (at edge or cloud) processes images from
remote IoT devices, the inference engine at the server may encounter various
quality levels of JPEG images as JPEG is the dominant image compression
standard. In this context, we evaluate \name with inputs of various JPEG quality
levels.
%in order to show that \name is not limited to a particular model, we also
We use the Cityscapes dataset, and build \name over \texttt{DRN-D-22}, a smaller
\texttt{DRN} model than the one in \S\ref{sec:res}, to show it is not limited to
a particular model.
%
%We prepare the JPEG-distorted datasets following \S\ref{sec:quality}. 
Following \S\ref{sec:quality}, we convert the dataset to 10 versions with
different JPEG quality levels and set the adaptor size as $0.2\%$ of the model
size (\S\ref{sec:dnns}), then the memory overhead is $0.2\%\times(10-1)=1.8\%$
over the \emph{original DNN}.  

%\begin{figure}[t]
%\begin{center}
%\includegraphics[width=0.99\linewidth]{figures/jpeg_drn22_accuracy}
%\end{center}
%\vspace{-5mm}
%        \caption{Comparison between \name and benchmark solutions over the Cityscapes dataset using DRN-D-22 model and JPEG quality levels.}
%	\label{fig:accuracy_jpeg_22}
%%\vspace{-5mm}
%\end{figure}

The results in Fig.~\ref{fig:accuracy_jpeg_22} confirm the high inference accuracy
of \name on all JPEG quality levels. In contrast, \emph{original DNN} suffers from
significant accuracy degradation at low quality levels, and its mIoU falls
below 40\% for quality under 50. \emph{Mixed training} also achieves lower accuracy 
than \name does, especially on inputs with high \& low qualities, because
it attempts to map different distorted versions of an image to the same inference 
output, which is difficult for a single DNN (Fig.~\ref{fig:no_single}).
\emph{DNN switching} yields accuracy similar to \name's, but at the cost of 
much higher memory cost (Table \ref{table:mem}), as it needs to keep multiple 
DNNs in the memory. 
When each quality level has the same probability to appear in the input, \name
achieves $18.12\%$ average accuracy gain over the \emph{original DNN} and $1.95\%$
gain over the \emph{mixed training}. The accuracy difference with the memory-hungry 
\emph{DNN switching} is only $0.22\%$.

%\newcolumntype{L}[1]{>{\centering\arraybackslash}p{#1}}
  \begin{table}[t]
\caption{Memory overhead.}
\footnotesize{
%\small{
\centering
\begin{tabular}{@{}p{0.118\textwidth}*{4}{L{\dimexpr0.088\textwidth-2\tabcolsep\relax}}@{}}
\toprule
& \multicolumn{2}{c}{\shortstack[l]{\emph{10 of JPEG
qualities}\\\texttt{(DRN-D-22)}}} & \multicolumn{2}{c}{\shortstack[l]{\emph{7 of
H.264 resolutions}\\\texttt{(DRN-D-38)}}} \\
\cmidrule(r{4pt}){2-3} 
\cmidrule(r{4pt}){4-5} 
& \multirow{2}{*}{\emph{Params}} &
\multirow{2}{*}{\shortstack[l]{\emph{Overhead}\\\emph{vs. original}}} & \multirow{2}{*}{\emph{Params}} &
\multirow{2}{*}{\shortstack[l]{\emph{Overhead}\\\emph{vs. original}}}  \\
& & & \\
\midrule
\textbf{\name}&16,193,465& 1.8\% &26,334,993& 1.2\%\\
\textbf{Original DNN}&15,907,139& 0\% &26,022,723& 0\%\\
\textbf{Stability training}&15,907,139& 0\% &26,022,723& 0\%\\
\textbf{Mixed training}&15,907,139& 0\%  &26,022,723& 0\%\\
\textbf{DNN switching}&159,071,390& 900\% &182,159,061& 600\%\\
\bottomrule
\end{tabular}
\label{table:mem}}
\end{table}

%
%For the DNN input distortion, we test 3 different types of distortions:
%\texttt{JPEG} image compression, \texttt{H.264} video encoding, and brightness
%change. For each type of distortion, we test 10 distortion levels as defined in
%\S\ref{sec:quality}. 
%
%Since \name is a generic adaptive inference architecture instead of a particular
%DNN model, we evaluate \name on top of two popular DNN models: the dilated
%residual networks (DRN)~\cite{drn1, drn2} and the Mask R-CNN
%model~\cite{mask_rcnn}. 
%%
%More specifically, we run DRN on the Cityscapes dataset~\cite{cityscapes} for
%urban driving scenery semantic segmentation and Mask R-CNN on the Penn-Fudan
%dataset~\cite{penn} for pedestrian segmentation and detection.
%

\subsection{H.264 video resolution}
\label{sec:res}
We then evaluate how \name guarantees high inference accuracy across dynamic
input video resolutions, by using the Cityscapes dataset, \texttt{DRN-D-38}
model, and the frame
scaling with 7 resolution levels as distortions (\S\ref{sec:quality}) 
We set the adaptor size to $0.2\%$ of the model size (\S\ref{sec:dnns}), then
the memory overhead is $0.2\%\times(7-1)=1.2\%$ over the \emph{original DNN}.  
%, following \S\ref{sec:quality}. 
%
%We
%use the \texttt{H.264} encoder to convert the image-based Cityscapes dataset to
%videos with different resolutions and have 7 different video resolutions. All
%these video frames have a CRF of 18 regardless of their resolutions.  
%
%\begin{figure}[t]
%\begin{center}
%\includegraphics[width=0.99\linewidth]{figures/res_h264_drn38}
%\end{center}
%\vspace{-5mm}
%   \caption{Comparison between \name and benchmarks for the
%different resolutions on the Cityscapes dataset.}
%\label{fig:res_h264}
%%\vspace{-5mm}
%\end{figure}
%
From the results in Fig.~\ref{fig:res_h264}, we see that both \name and
\emph{DNN switching} significantly outperform the \emph{original DNN} and
\emph{mixed training}, with around $4\%$ higher accuracy (mIoU) 
under resolutions lower than 1024\footnote{The DNN trained with original images 
achieves the highest accuracy at a smaller input size (Fig.~\ref{fig:res_h264}),
which implies that scaling the object to match field-of-view of the DNN kernel may 
compensate for the information loss. Nevertheless, this observation is out
of the scope of this paper.}. 
Although \name and \emph{DNN switching} enjoy similar accuracy, \name only costs
a $1.2\%$ memory overhead over the \emph{original DNN}
(Table~\ref{table:mem}), while \emph{DNN switching} requires keeping multiple DNNs
in the memory and thus suffers from a high memory overhead.
When each input resolution has an equal chance to appear, \name achieves $2.88\%$
higher average accuracy over the \emph{original DNN}, $1.70\%$ over the
\emph{mixed training}, and $0.71\%$ over \emph{DNN switching}.

\begin{figure*}[t]
    \centering
    \begin{minipage}[t]{0.32\linewidth}
	\includegraphics[width=\textwidth]{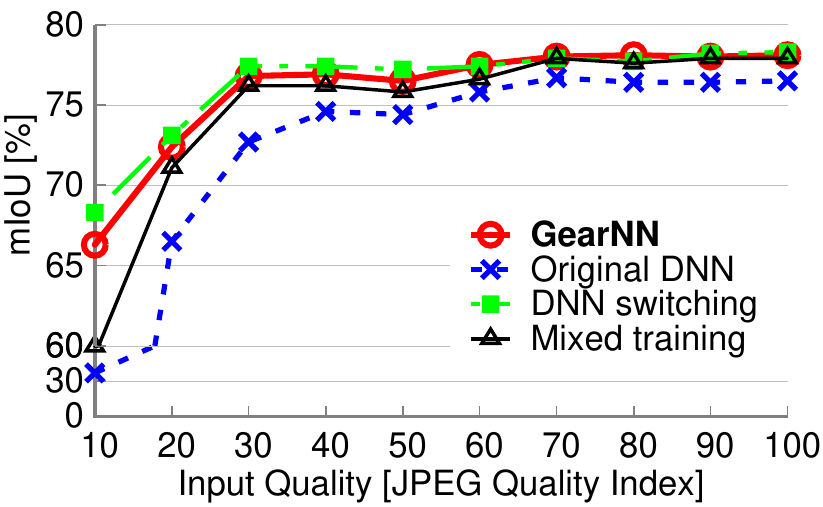}
	\vspace{-5mm}
	\caption{\name based on \texttt{Mask R-CNN} (semantic segmentation).}
	\label{fig:mask}
	\end{minipage}
	\hspace{3pt}
    \begin{minipage}[t]{0.32\linewidth}
	\includegraphics[width=\textwidth]{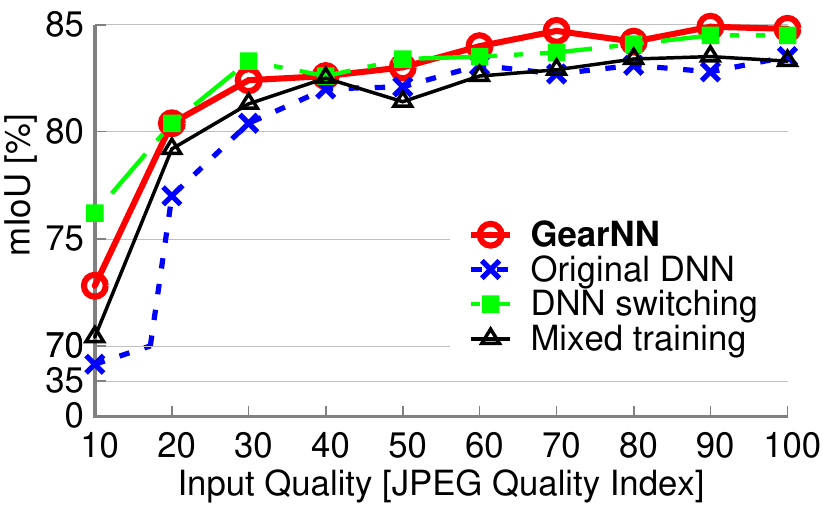}
	\vspace{-5mm}
	\caption{\name based on \texttt{Mask R-CNN} (bounding box detection)}
	\label{fig:mask_bbox}
	\end{minipage}
	\hspace{3pt}
    \begin{minipage}[t]{0.32\linewidth}
        \includegraphics[width=\textwidth]{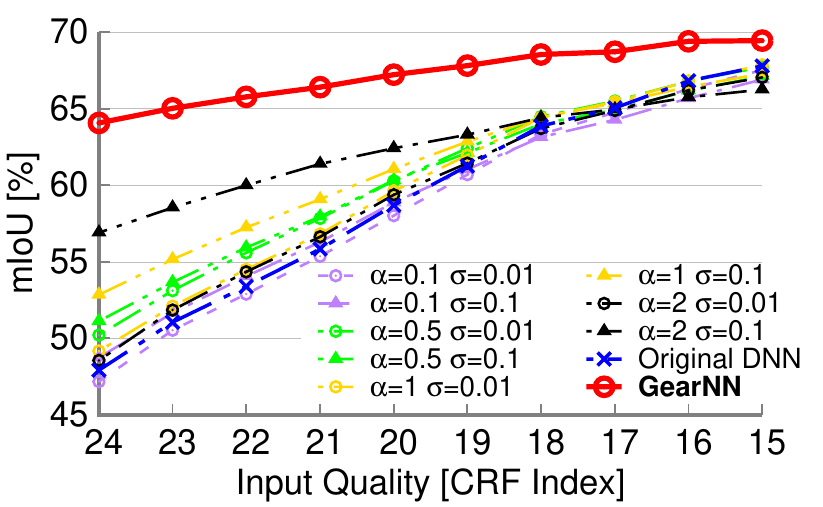}
        \vspace{-5mm}
        \caption{\name vs. \emph{stability training} with various configurations.}
	\label{fig:vs_stab}
    \end{minipage}
\end{figure*}

\subsection{Brightness level}
\label{sec:bright}
Another application scenario of \name is autonomous driving,
where the camera can capture video frames with diverse brightness levels. In
such a scenario, abrupt brightness change may cause even human eyes to 
malfunction temporarily. In this experiment, we reveal that the DNN-based computer 
vision also suffers from the same problem, and \name can address this by partially
adapting the DNN to the brightness changes. Following \S\ref{sec:quality}, 
we alter the brightness of the datasets and use the DNR-D-38 model.
%, and perform the training and evaluation using these datasets with various brightness levels.  

The results in Fig.~\ref{fig:bright_drn38} show the same performance
ranking as the experiments above -- \name and \emph{DNN switching} achieve comparable
accuracy, while \emph{mixed training} suffers significant accuracy loss 
on high or low input brightness. Next, the original DNN has the worst performance 
on all brightness levels. Overall, considering the 20 brightness levels from
0.1 to 2.0, \name costs merely $0.2\%\times20=4\%$ memory overhead, compared to 
the original DNN, while \emph{switching DNNs} costs $1900\%$ extra memory, 
rendering it impractical.
If all brightness levels have the same chance of appearing in the input, \name
achieves $5.11\%$ higher average accuracy than the \emph{original DNN}, $1.11\%$
than the \emph{mixed training}, and a slight $0.05\%$ accuracy difference over
the memory-hungry \emph{DNN switching}.

\subsection{Different datasets and tasks}
\label{sec:multi_datasets}
To demonstrate that \name is not bounded to a particular model or dataset, 
we run \name on pedestrian detection and segmentation with the Penn-Fudan dataset.
We use \texttt{Mask R-CNN} with \texttt{ResNet-50} as the backbone and 
perform both the segmentation and bounding box detection of the pedestrians. Our
\name prototype runs on top of the \texttt{Mask R-CNN} model and 
supports 10 JPEG input quality levels (\S\ref{sec:quality}), for both the pedestrian
segmentation (Fig.~\ref{fig:mask}) and bounding box detection
(Fig.~\ref{fig:mask_bbox}). Since we set the adaptor size to $0.4\%$ of the model
size (\S\ref{sec:dnns}), the memory overhead is $0.4\%\times(10-1)=3.6\%$
over the \emph{original DNN}.

From the segmentation results in Fig.~\ref{fig:mask}, we first see that the
\emph{original DNN} and \emph{mixed training} suffer from severe accuracy
loss when the input quality is below 20. 
\name has similar accuracy to \emph{DNN switching}, as in all other experiments. 
As the input quality increases, the accuracy of the \emph{original DNN} and
\emph{mixed training} remain below \name's. But the gap is smaller than
that of the Cityscapes dataset, because the PennFudan dataset has much lower
resolution (around 300$\sim$400 width \& height), containing less information than
Cityscapes (2048$\times$1024), and thus, the input distortions cause less
impact.
In Fig.~\ref{fig:mask_bbox}, we further plot the bounding box detection accuracy
measured from the same set of experiments. The results show that \name
always achieves higher accuracy than the \emph{original DNN} and \emph{mixed training},
especially at low input qualities. Meanwhile, it achieves similar
accuracy, while maintaining much lower memory overhead than \emph{DNN switching}.  

Overall, if all JPEG quality levels have the same probability to appear in the
input, \name achieves $5.16\%$ (segmentation) / $3.53\%$ (detection) higher
average accuracy than the \emph{original DNN}, $1.20\%$ (segmentation) / $1.33\%$
(detection) higher than the \emph{mixed training}, and a small $0.43\%$
(segmentation) / $0.24\%$ (detection) accuracy difference from the memory-hungry
\emph{DNN switching}. 

\subsection{Comparison with stability training}
\label{sec:stability}
There are only a few existing solutions in improving the DNN robustness to 
distorted/perturbed inputs. One of the most well-known works is the
\emph{stability training} proposed by Google. 
Therefore, we perform an extensive performance comparison against
\emph{stability training} under various configurations. 
In particular, we implement both \name and \emph{stability training} based on
the \texttt{DRN-D-38} model and perform the evaluation on the Cityscapes dataset
compressed by the H.264 encoder. For \emph{stability training}, there are two
configurable parameters: (i) $\alpha$ controls the tradeoff between the
stability and the accuracy on undistorted inputs, and (ii) $\sigma$ controls the
tolerance to distortions. We test multiple combinations of $\alpha$
and $\sigma$, then show the results in Fig.~\ref{fig:vs_stab}.

We first observe that \name outperforms all tested
combinations of $\alpha$ and $\sigma$ at any input distortion level. We also see
that $\alpha$ and $\sigma$ control the performance of \emph{stability training}
as expected. Overall, a higher $\alpha$ or $\sigma$ leads to better tolerance to
distortions. It essentially sacrifices the performance on the low-distortion
inputs to improve the performance on the high-distortion
inputs, so that the overall performance is balanced.  
Compared to the \emph{stability training} configuration with the highest
accuracy when $\alpha=2$ and $\sigma=0.1$, \name achieves $4.84\%$ average
accuracy gain, while the highest accuracy gain is $7.15\%$ at the CRF of 24 and
$3.18\%$ at the CRF of 15. 
In sum, \emph{stability training} uses a single DNN to fit a wide range of
distortions, and thus it has to tradeoff performance across different distortion
levels. In contrast, \name partially adapts the weights to fit the current
input distortion level, which guarantees its high accuracy and low memory
overhead.

%We also evaluate another task, bounding box detection, in
%Fig.~\ref{fig:mask_bbox}. In this test, we test 20 relative brightness levels
%levels from 0.2 to 2.0. We can observe that \name achieves much higher accuracy
%than the \emph{original DNN} and \emph{mixed training} when the
%visual input is too dark or too bright, especially when the relative brightness is below
%0.4 or above 1.6. Although \emph{DNN switching} has the
%highest accuracy in this case, its high memory overhead makes its deployment
%infeasible.

\section{Conclusion}
\label{sec:conclusion}
In this paper, we present \name, a memory-efficient adaptive DNN inference
architecture, to combat the diverse input distortions. \name enables robust
inference under a broad range of input distortions, without compromising memory
consumption. It identifies and adapts only the distortion-sensitive DNN
parameters, a tiny portion of the DNN (\eg, $0.2\%$ of the total size),
following the instantaneous input distortion level. 
Our evaluation demonstrates the superior performance of \name over benchmark
solutions such as \emph{stability training} from Google, when the distortion
level of input varies due to adaptive video resolution, JPEG compression, or
brightness change. As a general inference architecture, \name can be potentially
applied to many existing DNN models and enables them to accommodate the diverse
input distortions in the IoT era. 

%In this way,  can accommodate a diverse range of input distortion levels

%Further, \name achieves a higher accuracy on visual input with dynamic distortions 
%than a constant DNN with the same number of parameters. Meanwhile, it significantly
%reduces memory cost, compared to switching the entire DNN.

%The evaluation results showed that \name achieves comparable inference accuracy as switching the entire DNN while having tiny memory overhead compared to the original DNN.

%%%%%%%%% References
{\small
\bibliographystyle{ieee_fullname}
\bibliography{main}
}

\end{document}